\documentclass{svproc}

\usepackage[bookmarks=true]{hyperref}
\usepackage[numbers]{natbib}
\usepackage[bottom]{footmisc}

\usepackage{graphicx}
\usepackage{subfigure}
\graphicspath{{figures/}{./}} 
\usepackage{sidecap}

\usepackage[latin1]{inputenc}
\usepackage{fancyhdr}    
\usepackage{textcomp}  
\usepackage{amsmath} 
\usepackage{amssymb}
\usepackage{dsfont}
\usepackage{psfrag}
\usepackage{stfloats}
\usepackage{color}
\usepackage{multicol}
\usepackage{algorithmic}
\usepackage{algorithm}
\usepackage{array}
\usepackage{paralist}{}
\usepackage{wasysym}
\usepackage[normalem]{ulem}

\usepackage{anyfontsize}

\usepackage{url}

\begin{document}
\mainmatter 

\title{Redundant Robot Assignment on Graphs with Uncertain Edge Costs}
\titlerunning{Redundant Robot Assignment}

\author{Amanda Prorok}
\authorrunning{A. Prorok} 
\institute{Department of Computer Science \& Technology \\ University of Cambridge, UK \\ \email{asp45@cam.ac.uk}}

\maketitle

\begin{abstract} 
We provide a framework for the assignment of multiple robots to goal locations, when robot travel times are uncertain. 
Our premise is that \emph{time} is the most valuable asset in the system. Hence, we make use of \emph{redundant} robots to counter the effect of uncertainty and minimize the average waiting time at destinations. We apply our framework to transport networks represented as graphs, and consider uncertainty in the edge costs (i.e., travel time).
Since solving the redundant assignment problem is strongly NP-hard, we exploit structural properties of our problem to propose a polynomial-time solution with provable sub-optimality bounds. Our method uses \emph{distributive aggregate functions}, which allow us to efficiently (i.e., incrementally) compute the effective cost of assigning redundant robots. Experimental results on random graphs show that the deployment of redundant robots through our method reduces waiting times at goal locations, when edge traversals are uncertain.
\end{abstract}

\section{Introduction}
Technological advances are enabling the large-scale deployment of robots to solve various types of logistics problems, such as product delivery~\cite{grippa:2017}, warehousing~\cite{enright:2011}, and mobility-on-demand~\cite{pavone2012robotic}. 
Robot teams also hold the promise of delivering robust performance in unstructured or extreme environments~\cite{thayer:2001,kantor2003distributed}.
The commonality of many of these applications is that they require the assignment of robots to destinations.
%
In general, a solution to the assignment problem can be computed by a centralized unit that collects all robot-to-goal assignment costs (e.g., expected travel times) to determine the optimal assignment (e.g., by running the Hungarian algorithm). However, the optimality of this assignment hinges on the accuracy of the assignment cost estimates, i.e., the cost of traversing any path in the underlying representation. 
Unpredictable events along robot paths cause uncertain travel times. For example, in structured environments, we may encounter traffic accidents and congestion; in unstructured environments, we may encounter unsafe terrains and risky traversal conditions. 

We are interested in applications that require a fast arrival of robots at their destinations, but where knowledge about expected robot travel times may be imprecise or incomplete due to uncertain path conditions. 
Although robot assignments under random costs have gained a considerable amount of attention~\cite{krokhmal:2009,Nam:2017jwa,Nam:2015hz}, the focus has primarily been on providing analyses of the performance under noisy conditions. In this work, we propose a complementary method by making use of \emph{robot redundancy}.
In other words, the core idea of our work is to exploit {redundancy} to counter the adverse affect of uncertainty. 
Although the idea of engineering robust systems with {redundant resources} is not new in a broad sense~\cite{kulturel:2003,ghare:1969}, we are the first to consider redundant mechanisms for the problem of mobile robot assignment under uncertainty, with arbitrary and potentially correlated probability distributions. We believe that providing redundant robots will be a fundamental design feature for systems where time is the primary asset (e.g., rescue scenarios), and where an over-provisioning with respect to the number of robots is a minor concern.

\textbf{Background.} Our problem belongs to the class of Single-Task Robots, Multi-Robot Tasks (ST-MR)~\cite{Gerkey:2004il}, which considers the assignment of groups of robots that have a combined utility for any given task. The aim is to split the set of robots to form task-specific coalitions, such that the average task utilities over all coalition-to-task assignments is maximized (or, in our case, the average cost minimized). Formally, this can be cast as a set-partitioning problem, which is strongly NP-hard~\cite{garey:1978}. 
However, if we pay special attention to the objective function, the problem may reveal additional structure that can be exploited to find near-optimal solutions. In particular, for utility functions that satisfy a property of \emph{diminishing returns}, otherwise known as \emph{submodularity}, near-optimal approximations can be found~\cite{fujishige:2005}. 
{As we show in this work, and in a previous report~\cite{prorok:2018}, our approach provides diminishing returns in the number of redundant robots assigned to each goal}. This yields a \emph{supermodular} cost function, justifying the usage of efficient, greedy assignment algorithms. 

\textbf{Contributions.} In this paper, we solve the problem of selecting optimal redundant robot-to-goal matchings on transport graphs with uncertain edge costs. Our objective is to minimize the waiting time at the goals, whilst respecting an upper limit on how many (redundant) robots may be assigned. A previous report proved the supermodularity of our objective~\cite{prorok:2018}, and considered the special case where robot positions are uncertain, yet path costs are fixed. The present paper focuses on uncertain edge costs, which implies variances on path costs, and introduces the additional complexity of selecting among numerous partially correlated uncertain path choices. We show that we can capture these additional elements with a matroid constraint, and thus, can provide a polynomial-time algorithm that uses distributive aggregate functions to efficiently assemble the gain of assigning redundant robots. Finally, we provide evaluations on random graphs with uncertain, correlated edge costs.

\section{Problem Description}

We consider a system composed of $M$ goals and $N$ available robots. The size of the total robot deployment is constrained by $N_{\mathrm{d}}$, with $M \leq N_{\mathrm{d}} \leq N$.
This constraint is relevant for applications that run continuous assignments with fluctuating task demand, and enables the reservation of some robots for future deployments. Furthermore, it allows operators to limit the cost of redundant robot deployments, for example by monitoring and capping energy consumption. 
Our problem considers the assignment of robots to goals via a path, where, for each robot-to-goal assignment, multiple possible paths with random costs exist. Without loss of generality, we assume that $K$ possible paths exist between each robot and each goal.
We seek to find a minimum cost matching, such that all goals are covered, and any goal may be assigned multiple robots, while respecting the limit on the maximum deployment size $N_{\mathrm{d}}$. 

\textbf{Assignment with Random Path Costs.} 
Consider a graph $\mathcal{B}= (\mathcal{U},\mathcal{F}, \mathcal{C})$.
The set of vertices $\mathcal{U}$ is partitioned into two subsets $\mathcal{U}_r$ and $\mathcal{U}_g$, such that $\mathcal{U}_r = \bigcup_i r_i,~i=1,\ldots,N$ contains all robot nodes, $\mathcal{U}_g = \bigcup_j g_j, ~j=1,\ldots, M$ contains all goal nodes, and $\mathcal{U} = \mathcal{U}_r \, \cup \, \mathcal{U}_g$, $\mathcal{U}_r \, \cap \, \mathcal{U}_g = \emptyset$. We define a fixed  number $K$ of possible path routes that lead any robot to any goal. The edge set $\mathcal{F} = \{(i,j,k)| i \in \mathcal{U}_r, j \in \mathcal{U}_g, k \in 1,\ldots,K, \forall \, i,j,k\}$ is complete, meaning that any robot can reach any goal node, and that up to $K$ possible path choices exist for any pair $(i,j)$.
The tuple $(i,j,k)$ indicates that robot $i$ is assigned to goal $j$ through path $k$.
Since the travel time for a robot to reach its goal is uncertain, we represent the weight of each edge $(i,j,k)$ by a random variable $C_{ijk} \in \mathcal{C}$, where $\mathcal{C}$ is a set of random variables. The set $\mathcal{C}$ has a joint distribution $\mathcal{D}$.
Hence, $C_{ijk}$ can be arbitrarily defined for any edge; in particular, edge costs may be correlated, and we do not make use of \emph{i.i.d.} assumptions.

\begin{SCfigure}[][tb]
\centering
\psfrag{A}[cc][][0.8]{$\mathcal{A}$}
\psfrag{O}[cc][][0.8]{$\mathcal{O}$}
\psfrag{i}[cc][][0.8]{$i$}
\psfrag{j}[cc][][0.8]{$j$}
\psfrag{1}[cc][][0.8]{$r_1$}
\psfrag{2}[cc][][0.8]{$r_2$}
\psfrag{3}[cc][][0.8]{$r_3$}
\psfrag{4}[cc][][0.8]{$r_4$}
\psfrag{5}[cc][][0.8]{$r_5$}
\psfrag{6}[cc][][0.8]{$r_6$}
\psfrag{7}[cc][][0.8]{$g_1$}
\psfrag{8}[cc][][0.8]{$g_2$}
\psfrag{a}[cc][][0.8]{$\mathcal{U}_r$}
\psfrag{b}[cc][][0.8]{$\mathcal{U}_g$}
\psfrag{c}[cc][][0.8]{${C}_{621}$}
\psfrag{D}[cc][][0.8]{${C}_{622}$}
\includegraphics[width=0.44\columnwidth]{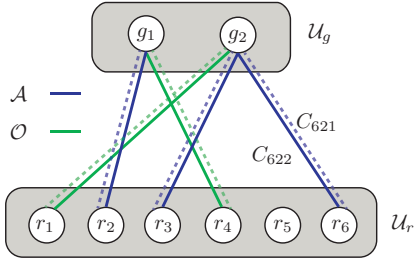}
\caption{Sketch of redundant robot-to-goal matching, for a total of $N=6$ available robots, $M=2$ goals, $K=2$ path options, and a deployment size of $N_{\mathrm{d}}=5$. Edges in $\mathcal{O}$ represent the initial assignment, and edges in $\mathcal{A}$ represent the redundant assignment. The random variable $C_{ijk}$ represents the uncertain travel time it takes robot $i$ to reach goal $j$ via path $k$.
\label{fig:matching}}
\end{SCfigure}

We are interested in settings where an initial non-redundant assignment has been made. We consider an initial assignment $\mathcal{O} \subset \mathcal{F}$, such that $\forall j |\{i | (i,j,k) \in \mathcal{O} \}| = 1$, $\forall i |\{j | (i,j,k) \in \mathcal{O} \}| \leq 1$, and $\forall i, j |\{k | (i,j,k) \in \mathcal{O} \}| \leq 1$. In other words, $\mathcal{O}$ covers every goal with one robot, and any robot is assigned to at most one goal through one given {path}. 
Given an initial assignment, the aim of this paper is to find an optimal set of assignments $\mathcal{A}^\star \subset \mathcal{F} \setminus \mathcal{O}$ for the remaining $N_{\mathrm{d}} - M$ robots~\footnote{Without $\mathcal{O}$, any solution that is smaller in size than $M$ would lead to an infinite waiting time, and hence, the objective function looses its supermodular property. The assumption that we already have an initial assignment is necessary, for the developments that follow.}. From here on, we denote $\mathcal{F} \setminus \mathcal{O}$ by $\mathcal{F}_\mathcal{O}$.
Fig.~\ref{fig:matching} illustrates a simple robot-to-goal matching.

\textbf{Optimization Problem.}
The main novelty of our approach is the use of redundant robots to help counter the adverse effect of uncertainty.
If the system admits a sufficiently large number of robots (i.e., $N_{\mathrm{d}} - M > 0$), we can assign multiple robots to the same goal, while still ensuring that all goals are assigned at least one robot. 
A key consideration is that the performance at each goal is measured by an \emph{aggregate cost} function that considers the joint performance of all assigned robots at that goal.

\begin{definition}[Aggregate Cost] 
\label{def:aggregate}
We define an aggregate function $\Lambda: 2^{\mathcal{F}} \mapsto \mathbb{R}$ that operates over the set of edges incident to a given node, and returns a scalar that represents the aggregate cost over the weights of these edges. If $I_j(\mathcal{A})$ is the set of incident edges to node $j$ in the set of edges $\mathcal{A}$, and is equal to $\{(i,j,k) | \forall i, k \,\mathrm{with}\, (i,j,k) \in \mathcal{A}\}$, then we can write the aggregate cost for goal $j$ as 
\begin{equation}
\Lambda(I_j(\mathcal{A})). 
\end{equation}
\end{definition}
This definition allows us to formulate our objective function. For a given set $\mathcal{A}$ of edges that define robot to goal assignments, we wish to measure the average aggregate cost over all goals, in expectation over the random edge costs:
\begin{equation} \label{eq:objective}
J_{\mathcal{O}}(\mathcal{A}) = \frac{1}{M} \, \sum_{j=1}^{M} ~\mathop{\mathbb{E}}\limits_{\mathcal{C}}\,\left[ \Lambda(I_j(\mathcal{A \cup \mathcal{O}}))  \right].
\end{equation}
We note that when no redundant robots are deployed, the assignment is reduced to the set $\mathcal{O}$, for which the performance is measured as
\begin{equation} \label{eq:J0}
J_0 = J_{\mathcal{O}}(\emptyset) = \frac{1}{M} \, \sum_{j=1}^{M} ~\mathop{\mathbb{E}}\limits_{\mathcal{C}}\,[C_{ijk} | (i,j,k) \in \mathcal{O}]. 
\end{equation}
We formalize our problem as follows.
\begin{problem}\label{prob:problem1} \emph{Optimal matching of redundant robots under cardinality constraints and with multiple path choices:}
Given $N$ available robots, $M$ goals with uncertain robot-to-goal assignment costs, $K$ possible paths from each robot to each goal, and an initial assignment $\mathcal{O}$, find a matching $\mathcal{A} \subseteq \mathcal{F}_{\mathcal{O}}$ of redundant robots to goals such that the average cost over all goals is minimized, and the total number of robots deployed is $N_{\mathrm{d}}$. This is formally stated as:
\begin{eqnarray}
    \label{eq:opt_prob1}
    \mathop{\mathrm{argmin}}\limits_{\mathcal{A} \subseteq \mathcal{F}_{\mathcal{O}}} && J_{\mathcal{O}}(\mathcal{A}) \label{eq:objective1}\\
    \mathrm{subject~to}    
     &&   \forall i |\{ j | (i,j,k) \in \mathcal{A} \cup \mathcal{O} \} | \leq 1 \label{eq:constraint1}\\
     &&    |\mathcal{A}| = N_{\mathrm{d}} - M \label{eq:constraint2}
     \end{eqnarray}
\end{problem}

\begin{figure}[t]
\centering
\psfrag{b}[cc][][0.7]{path B}
\psfrag{a}[cc][][0.7]{path A}
\psfrag{c}[cc][][0.7]{path C}
\psfrag{v}[cc][][0.8]{$\mathcal{R}_1$}
\psfrag{x}[cc][][0.8]{$\mathcal{R}_2$}
\psfrag{g}[cc][][0.8]{Goal}
\psfrag{i}[cc][][0.7][90]{PDF}
\psfrag{e}[cc][][0.7]{Travel Time [s]}
\subfigure{\includegraphics[width=0.33\columnwidth]{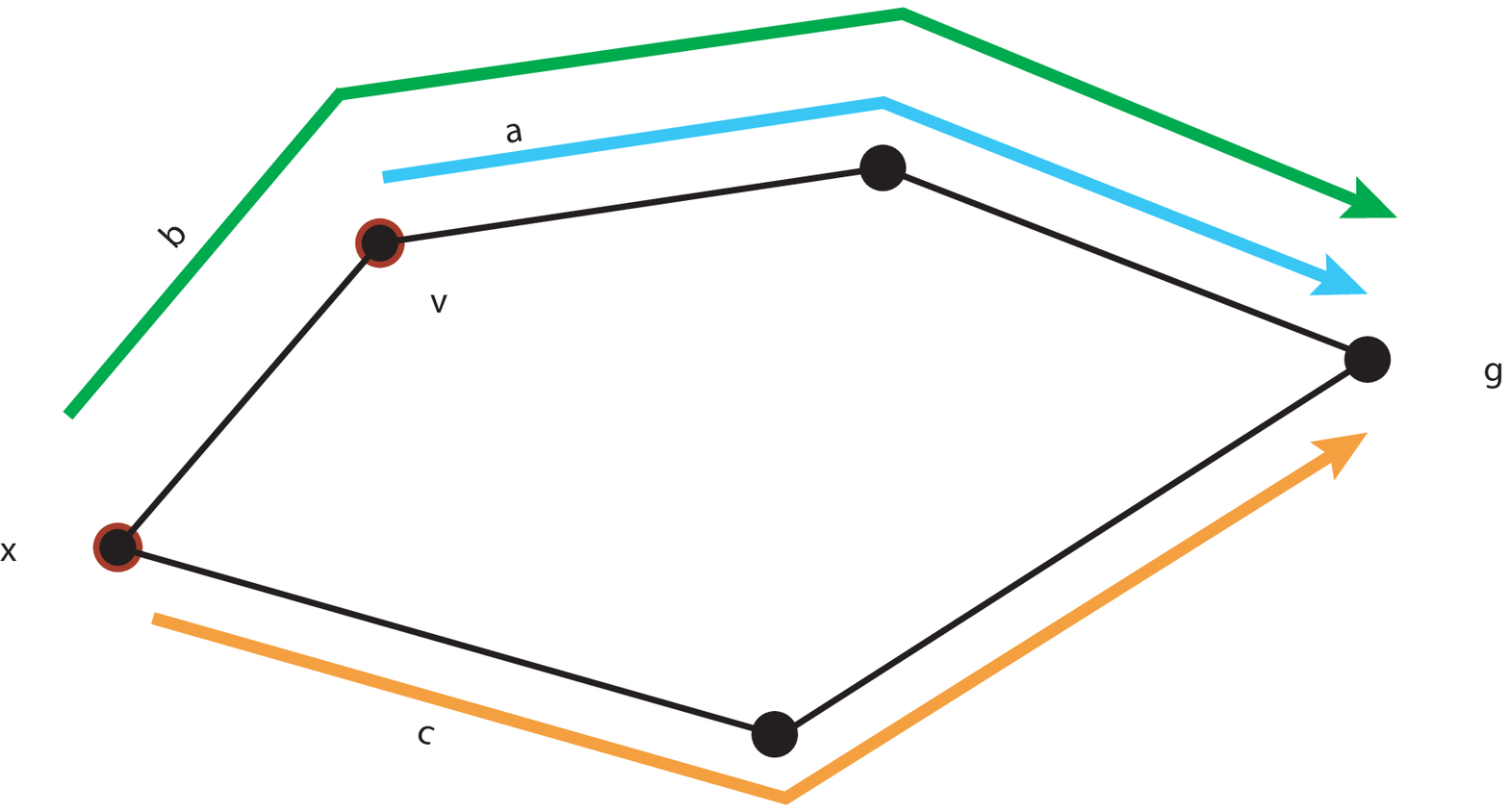}}\hspace{0.7cm}
\subfigure{\includegraphics[width=0.44\columnwidth]{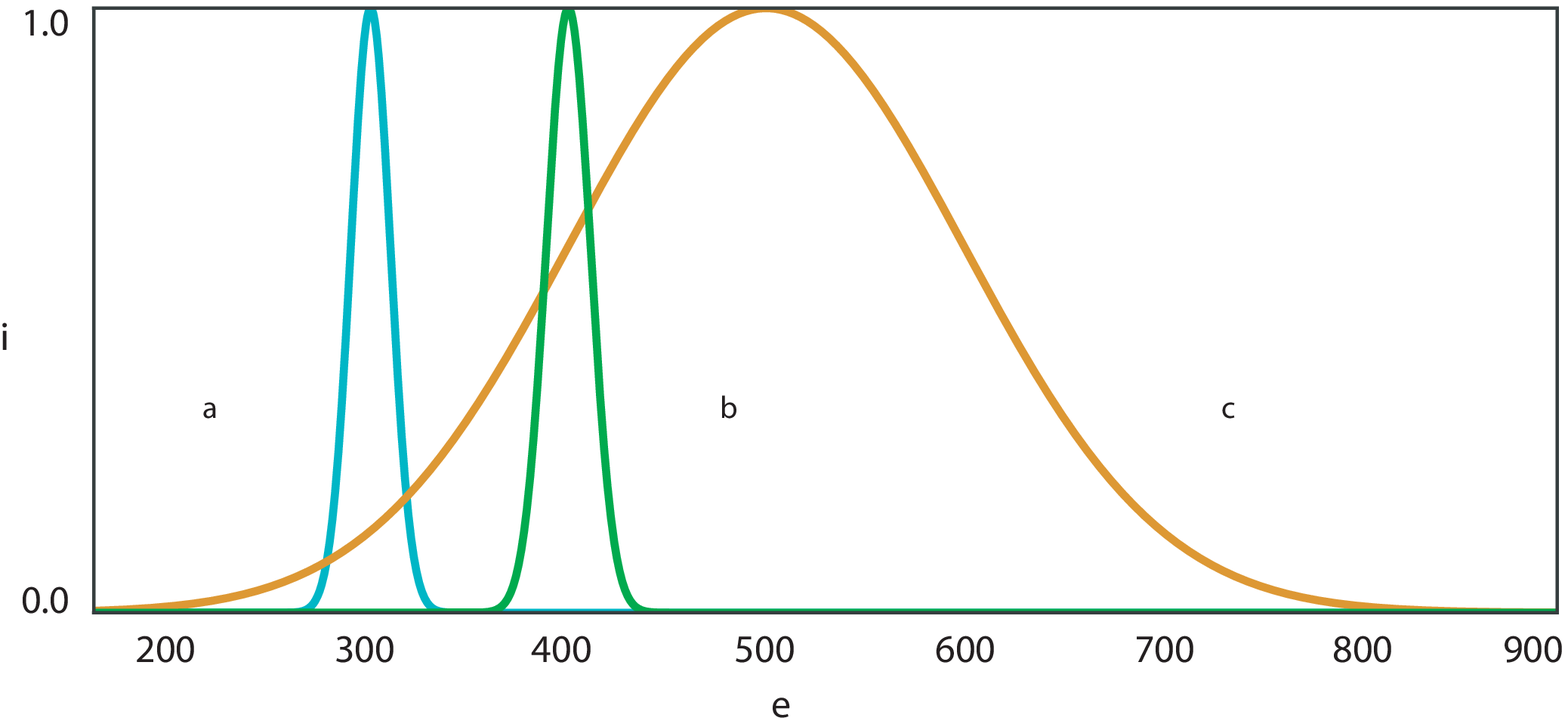}}
\caption{Example scenario. Robot $\mathcal{R}_1$ is assigned to the goal via path A. Robot $\mathcal{R}_2$ is a redundant robot, and must choose between paths B and C.
\label{fig:graph_sketch}}
\end{figure}

\begin{figure}[t]
\centering
\psfrag{a}[cc][][0.8][90]{Travel time along path \bf{B}}
\psfrag{x}[cc][][0.8]{Travel time along path \bf{A}}
\psfrag{c}[cc][][0.8][90]{Travel time along path \bf{C}}
\subfigure[]{\includegraphics[width=0.41\columnwidth]{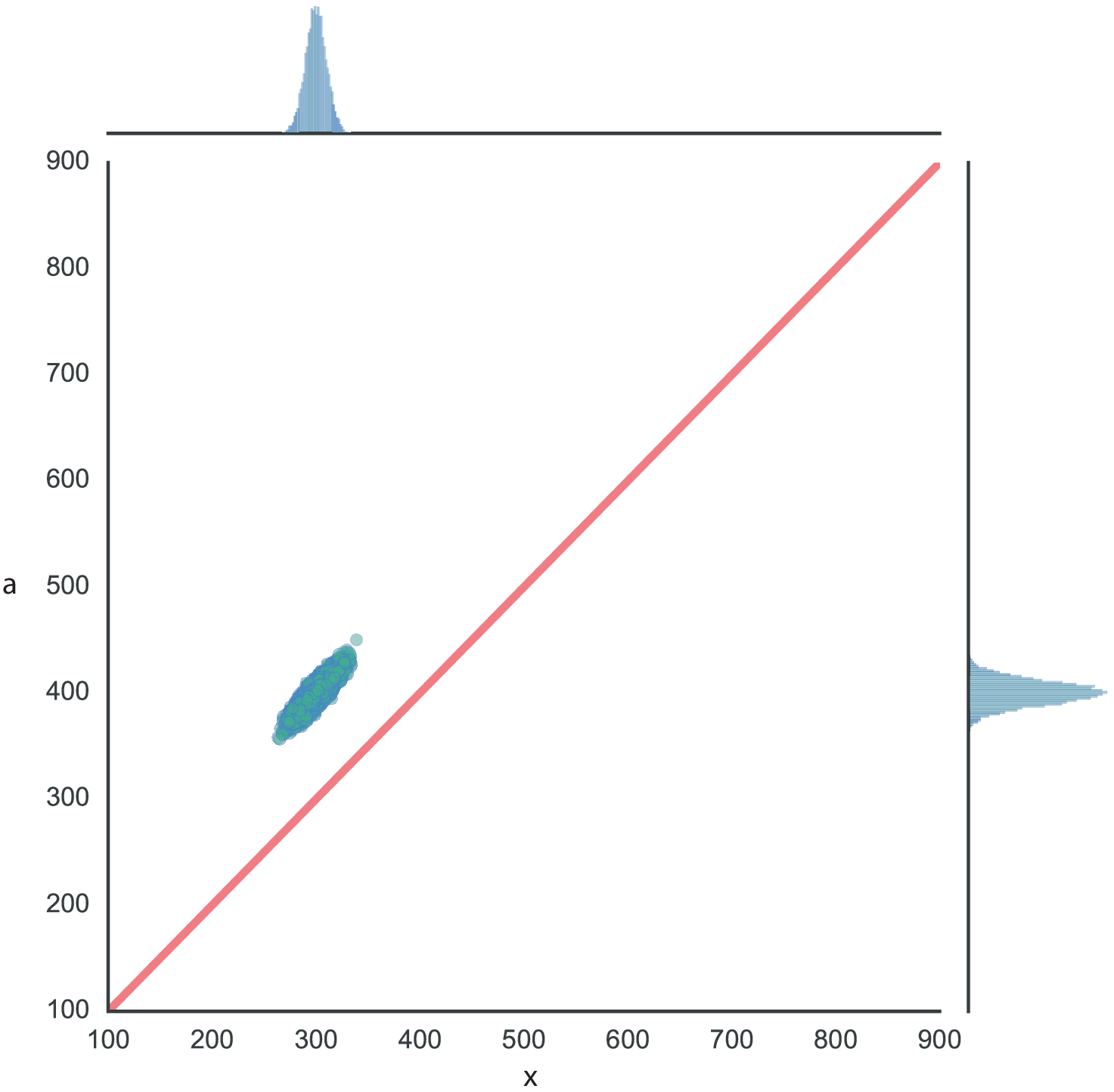}}\hspace{0.3cm}
\subfigure[]{\includegraphics[width=0.41\columnwidth]{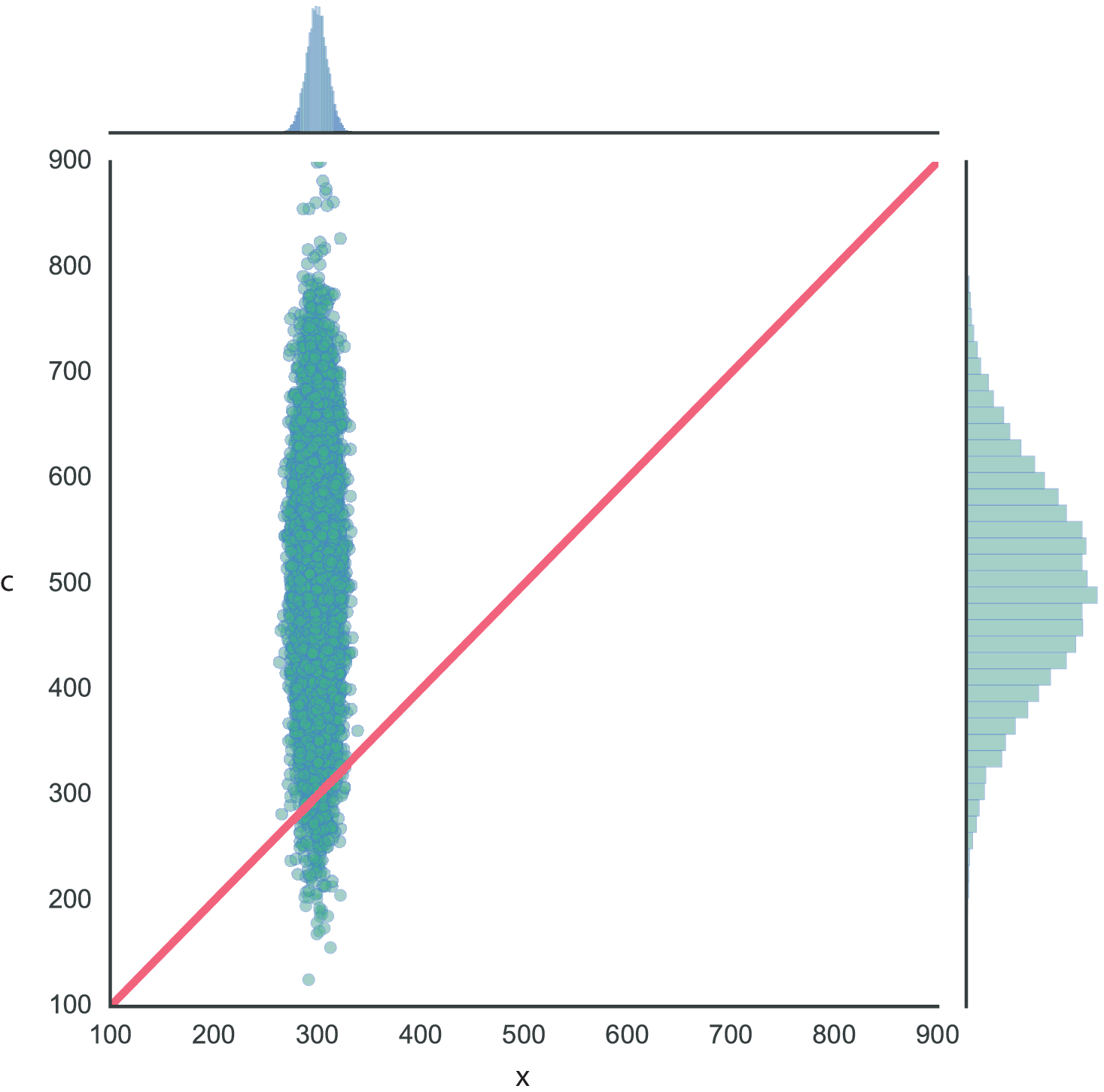}}
\caption{Joint distributions of travel times, for (a) paths A and B, and (b) paths A and C. The Pearson correlation coefficient is (a) 0.89 and (b) 0.0. The red line shows axis equality (i.e., equal travel times). If robot $\mathcal{R}_2$ chooses path C, there is a small chance that the waiting time at the goal will be improved upon, despite being slower than path B on average.
\label{fig:corr_gaussians}}
\end{figure}

Fig.~\ref{fig:graph_sketch} shows an example of an assignment problem with multiple path choices. In this particular example, robot $\mathcal{R}_1$ has already been assigned to the goal and $\mathcal{R}_2$ must choose between paths B and C. Path B is correlated with path A (they share two edges). Path C appears to take longer, and the intuitive choice would be path B. However, as Fig.~\ref{fig:corr_gaussians} shows, there is a small chance that path C leads to an improved waiting time at the goal, which cannot occur would robot $\mathcal{R}_2$ choose path B. Hence, the goal of our method is to identify combinations of robot assignments and paths that lead to the highest possible performance improvement. In the experimental section (Sec.~\ref{sec:eval}) we show how this leads to a notion of robot complementarity and diversity.
%

\section{Preliminaries}
\label{sec:background}

Our method is underpinned by the following key insight: as we assign additional robots to a given goal, the assignment cost for that goal decreases by a diminishing amount. This property is known as \emph{supermodularity}. In addition, the cardinality constraint imposed by our maximum deployment size $N_{\mathrm{d}}$ can be represented by a \emph{matroid} that also respects multiple path options.
These two properties aid us in finding an efficient algorithm to solve the redundant assignment problem, as elaborated later in Sec.~\ref{sec:approach}. The following definitions introduce the underlying concepts. 

\begin{definition}[Marginal decrease]
For a finite set $\mathcal{F}$ and a given set function $J: 2^{\mathcal{F}} \mapsto \mathbb{R}$, the marginal decrease of $J$ at a subset $\mathcal{A} \subseteq \mathcal{F}$ with respect to an element $x \in \mathcal{F} \setminus \mathcal{A}$ is:
\begin{eqnarray}
\Delta_{J}(x|\mathcal{A}) \triangleq J(\mathcal{A}) - J(\mathcal{A} \cup \{x\}).
\end{eqnarray}
\end{definition}

\begin{definition}[Supermodular function]\label{def:supermodular}
Let $J: 2^{\mathcal{F}} \mapsto \mathbb{R}$ and $\mathcal{A} \subseteq \mathcal{B} \subseteq \mathcal{F}$. The set function $J$ is supermodular if and only if for any $x \in \mathcal{F} \setminus \mathcal{B}$:
\begin{eqnarray}
\Delta_{J}(x| \mathcal{A}) \geq \Delta_{J}(x| \mathcal{B})
\end{eqnarray}
\end{definition}
The definition implies that adding an element $x$ to a set $\mathcal{A}$ results in a larger marginal decrease than when $x$ is added to a superset of $\mathcal{A}$. This property is known as a property of diminishing returns from an added element $x$ as the set it is added to grows larger. 
\begin{remark}\label{remark:submodular}
A function $J$ is submodular if $-J$ is supermodular. 
\end{remark}

\begin{definition}[Matroid] Given a finite ground set $\mathcal{F}$ and $\mathcal{I} \subseteq 2^{\mathcal{F}}$ a family of subsets of $\mathcal{F}$, an independence system is an ordered pair $(\mathcal{F}, \mathcal{I})$ with the following two properties: {(i)} $\emptyset \in \mathcal{I}$, and {(ii)} for every $\mathcal{A} \in \mathcal{I}$, with $\mathcal{B} \subseteq \mathcal{A}$, implies that $\mathcal{B} \in \mathcal{I}$.
The second property is known as downwards-closed --- in other words, every subset of an independent set is independent. 
An independence system $(\mathcal{F}, \mathcal{I})$ is a matroid if it also satisfies the augmentation property, that is, for every $\mathcal{A}, \mathcal{B} \in \mathcal{I}$, with $|\mathcal{A}| > |\mathcal{B}|$, there exists an element $a \in \mathcal{A} \setminus \mathcal{B}$ such that $\{a\} \cup \mathcal{B} \in \mathcal{I}$.
\end{definition}

For an optimization problem $\min J(\mathcal{A})$ such that $\mathcal{A} \subset \mathcal{I}$, where $J$ is supermodular and $\mathcal{I}$ is an independence system, we can apply a \emph{greedy} approximation algorithm.
This approach works as follows.
At each iteration $k$, an element $a$ is added to the solution set $\mathcal{A}_{k-1}$ such that it maximizes the marginal decrease given this current set,
\begin{equation}
\label{eq:greedy_max}
a \gets \mathrm{argmax}_{ x \in \mathcal{F}_E (\mathcal{A}_{k-1}, \mathcal{I}) } \Delta_J (x|\mathcal{A}_{k-1})
\end{equation}
where $\mathcal{F}_{E}$ denotes the set of eligible elements, given the independence system $\mathcal{I}$ and ground set $\mathcal{F}$, defined as
\begin{equation}
\label{eq:greedy_set}
\mathcal{F}_{E}(\mathcal{A}_{k}, \mathcal{I}) \triangleq \{ x \in \mathcal{F} \setminus \mathcal{A}_k \,|\, \mathcal{A}_k \cup \{ x\} \in \mathcal{I} \}.
\end{equation} 
A key property is that optimization with this algorithm yields a $1/2$ approximation ratio~\cite{fisher:1978}.
In our case, this is equivalent to \emph{Greedy} returning a set $\mathcal{A}^\star$ with ratio $J(\mathcal{A}^\star) \leq \frac{1}{2} \, (J^\star + J_0)$ where $J^\star$ is the optimal cost, and $J_0$ is the system's baseline performance (without redundant assignments)~\cite{prorok:2018}.
For matroid rank $r$ and ground set size $n$, \emph{Greedy} requires $O(nr)$ calls to the objective function.

\section{Algorithmic Approach}
\label{sec:approach}
The section above establishes that if we are given a supermodular objective function that satisfies a matroid constraint, we can employ a greedy algorithm to solve our problem within known optimality bounds. However, to maintain the efficiency of such a greedy assignment algorithm, we need to ensure that the evaluation of the objective function itself is efficient (and can be computed in polynomial time). Towards this end, we develop a dynamic programming approach that hinges on a definition of \emph{incrementally computable functions}, which we apply to our redundant assignment problem. The following paragraphs elaborate our methodology --- first, in Sec.~\ref{sec:matroid}, we show that the matroid constraint applies to our problem setting, and second, in Sec.~\ref{sec:dp_greedy}, we show how Greedy is implemented efficiently through a dynamic programming approach. The resulting routine is shown in Algorithm~\ref{alg1}.

\subsection{Matroid Constraint}
\label{sec:matroid}
In the following we show that the problem of assigning redundant robots with multiple path options satisfies the properties of a matroid. Following constraints~\eqref{eq:constraint1} and~\eqref{eq:constraint2}, our problem considers the matroid $(\mathcal{F}_\mathcal{O}, \mathcal{I}_\mathcal{O})$, with
\begin{eqnarray}
\mathcal{I}_\mathcal{O} \triangleq && \{ \mathcal{A} | \mathcal{A} \subseteq \mathcal{F}_\mathcal{O} \,\land\, 
|\mathcal{A}| \leq N_{\mathrm{d}}-M  \,\land\,
\forall i |\{j|(i,j,k)\in \mathcal{A} \cup \mathcal{O}\}| \leq1  \}. \label{eq:IO}
\end{eqnarray}
By the definition of a matroid, any valid assignment must be an element of the family of independent sets $\mathcal{I}_\mathcal{O}$.
Firstly, the empty set is a valid solution, in which case our objective function is reduced to $J_0$, as given by~\eqref{eq:J0}. Secondly, our system is downwards-closed: for any valid robot-to-goal assignment $\mathcal{A} \in \mathcal{I}_\mathcal{O}$, any subset of assignments $\mathcal{B} \subseteq \mathcal{A}$ is also a valid assignment by~\eqref{eq:constraint1}. Thirdly, we can show that our system satisfies the augmentation property. For any two valid assignments $\mathcal{A}$ and $\mathcal{B}$, $|\mathcal{A}| > |\mathcal{B}|$ implies that there is at least one robot assigned to a goal in set $\mathcal{A}$ that is unassigned in set $\mathcal{B}$, irrespective of what path was selected. Hence, adding that robot-to-goal assignment to set $\mathcal{B}$ still satisfies~\eqref{eq:constraint1} and maintains the validity of the solution. We note that the augmentation property implies that all maximal solution sets have the same cardinality $N_{\mathrm{d}} - M$, which corresponds to the \emph{rank} of our matroid. 

\subsection{Greedy Assignment with Dynamic Programming}
\label{sec:dp_greedy}
Our work considers uncertainty models, represented by arbitrary distributions that are also capable of capturing correlations between random variables. Our approach is to consider a sampling-based method that takes $S$ samples from the $MNK$-dimensional joint distribution $\mathcal{D}$~\footnote{We note that if an analytical model is known, this can be used instead.}. 
Our aim is to ensure that the computation of the aggregate cost $\Lambda$ that assembles the performances of the redundant robots (see Def.~\ref{def:aggregate}) does not incur additional complexity that depends on the number of robots $N$, the deployment size $N_{\mathrm{d}}$, or number of tasks $M$. 
Our insight is that, in a number of practical cases, $\Lambda$ is a \emph{distributive aggregate function} and is {incrementally computable}~\cite{palpanas:2002}.
This allows us to implement a dynamic programming approach, as shown in Algorithm~\ref{alg1}.

\begin{definition}[Distributive Aggregate Function]
We define a class $\Delta$ of \uline{distributive} aggregate functions $\delta: A \mapsto \mathbb{R}$, for $A \subset \mathbb{R}$, such that $\delta \in \Delta$ only if $\delta(A \cup x)$ can be computed incrementally, as a function of the old value $\delta(A)$ and new value $x$ only.
\end{definition}
Sec.~\ref{sec:application} gives an implementation of distributive aggregate functions for the specific application considered in this work.

\begin{proposition}\label{theorem:alg_greedy}
Algorithm~\ref{alg1} is a valid instantiation of Greedy, and has complexity $O((N_{\mathrm{d}}-M) N M K S)$, if \emph{(i)} $\Lambda$ is a distributive aggregate function, \emph{(ii)} $J_{\mathcal{O}}$ is supermodular, and \emph{(iii)} $(F_{\mathcal{O}}, \mathcal{I}_\mathcal{O})$ is a matroid constraint.
\end{proposition}
\begin{algorithm}[tb]
\caption{{\small Greedy Redundant Assignment with Dynamic Programming}}
\label{alg1}
\begin{algorithmic}[1]
\REQUIRE Graph $\mathcal{B} = (\mathcal{U},\mathcal{F}, \mathcal{C})$, size of deployment $N_{\mathrm{d}}$, initial assignment $\mathcal{O}$
\ENSURE Set of edges $\mathcal{A}$ defining redundant assignments
\STATE $\mathcal{A} \gets \emptyset$
\STATE $\mathcal{F}_{\mathcal{O}} \gets \mathcal{F} \setminus \mathcal{O} $
\STATE $\mathcal{I}_{\mathcal{O}} \gets $ Eq.~\eqref{eq:IO}
\STATE $\hat{\mathcal{C}} \gets \mathrm{sample}~S~\mathrm{samples~from~} MNK\mathrm{-dimensional~distribution}~\mathcal{D}$
\FOR {$\hat{C}_{ijk} \in \hat{\mathcal{C}}$}
\STATE $\texttt{samples}[(i,j,k)] \gets \hat{C}_{ijk}$
\ENDFOR
\FOR {$(i,j,k) \in \mathcal{O}$}
\STATE $\texttt{task\_state}[j] \gets \texttt{samples}[(i,j,k)]$
\ENDFOR
\FOR{$d \in \{1,\ldots,N_{\mathrm{d}}-M\}$}
\STATE $\Delta_{J_{\mathcal{O}}}^{\star} \gets -\infty$
\STATE  $\mathcal{F}_{\mathcal{O},E} \gets \{(i,j,k) \in \mathcal{F}_{\mathcal{O}} \setminus \mathcal{A} \,| \, \mathcal{A} \cup \{(i,j,k)\} \in \mathcal{I}_{\mathcal{O}} \}$
\FOR{ $(i,j,k) \in \mathcal{F}_{\mathcal{O},E}$}
\STATE $\texttt{curr\_state} \gets \frac{1}{S} \sum_{z=1}^S \texttt{task\_state}[j]_z$
\STATE $\texttt{new\_state} \gets \frac{1}{S} \sum_{z=1}^S \Lambda(\texttt{task\_state}[j]_z, \texttt{samples}[(i,j,k)]_z)$
\STATE $\Delta_{J_{\mathcal{O}}} \gets \texttt{curr\_state} - \texttt{new\_state}$
\IF {$\Delta_{J_{\mathcal{O}}} > \Delta_{J_{\mathcal{O}}}^\star$}
\STATE {$\Delta_{J_{\mathcal{O}}}^\star \gets \Delta_{J_{\mathcal{O}}}$}
\STATE $(i^\star,j^\star,k^\star) \gets (i,j,k)$
\ENDIF
\ENDFOR
\STATE $\mathcal{A} \gets \mathcal{A} \cup (i^\star,j^\star,k^\star)$
\STATE $\texttt{task\_state}[j] \gets \Lambda.(\texttt{task\_state}[j], \texttt{samples}[(i^\star,j^\star,k^\star)])$ // element-wise $\Lambda$
\ENDFOR
\RETURN $\mathcal{A}$
\end{algorithmic}
\end{algorithm}

Algorithm~\ref{alg1} works as follows. Lines 1-10 initialize the data structures. 
In particular, lines 4-7 pre-sample a fixed set of $S$ samples (which amounts to sampling $S$ values for each of the $MNK$ edges). Pre-sampling allows the algorithm to maintain the supermodularity property.
For the remaining number of robots to be deployed, we proceed with a greedy assignment. Line 13 constructs the set of eligible assignments, as in Eq.~\eqref{eq:greedy_set}. Then, for all eligible assignment candidates, we compute the marginal cost decrease incurred by adding that assignment to goal $j$. In order to do this, Line 16 computes the new aggregate cost function. This is done incrementally, since $\Lambda$ is a distributive aggregate function. Overall, Lines 14-22 are equivalent to Eq.~\eqref{eq:greedy_max}, with Lines 18-21 retaining the best assignment candidate. Line 23 adds the best candidate to the current solution, and Line 24 updates the aggregate cost incurred at the goal the new robot was assigned to. 
Our approach requires $O((N_{\mathrm{d}}-M) N M K)$ calls to the objective function, and the objective function is computed in $O(S)$.

\section{Application to Transport Networks}
\label{sec:application}
We are interested in applications that use graphs to represent possible robot routes (from origins to goals), where the cost of traversing any individual route (or edge) is \emph{uncertain}.
We represent routes via a weighted directed graph, $\mathcal{G}= (\mathcal{V},\mathcal{E}, \mathcal{W})$. Vertices in the set $\mathcal{V}$ represent geographic locations. Nodes $u$ and $v$ are connected by an edge if $(u, v) \in \mathcal{E}$. We assume the graph $\mathcal{G}$ is a strongly connected graph, i.e., a path exists between any pair of vertices. The $k$th path between an origin node $i$ and goal node $j$ is denoted by tuple $(i,j,k)$. Paths between a same origin $i$ and goal $j$ are distinct if there is at least one edge in one path that is not present in the other path. 
The random variable $C_{ijk}$ captures the estimated travel time for robot $i$ to reach a goal $j$ via path $k$. We define a random variable $w_{uv} \in \mathcal{W}$ that represents the time needed to traverse an edge $(u,v)$. We formulate the estimated travel time on a path as:
\begin{equation}
\label{eq:cost}
C_{ijk}= \sum_{(u,v) \in \mathcal{S}_{ijk}} w_{uv}, \mathrm{~~and~thus~~}
\mathbb{E}[C_{ijk}]= \sum_{(u,v) \in \mathcal{S}_{ijk}} \mathbb{E}[w_{uv}],
\end{equation}
where $\mathcal{S}_{ijk}$ is the set of edges on path $k$ between node $i$ and node $j$. 

Let us consider a {\emph{`first-come, first-to-serve'}} principle, by which only the fastest robot to reach a task actually services it. Redundancy, as defined in Def.~\ref{def:aggregate}, allows us to reduce the waiting time at the goal locations: when multiple robots travel to the same destination, only the travel time of the fastest robot counts. This is further specified in the following definition.
\begin{definition}[Effective Waiting Time]
\label{def:waiting_time}
Since only the first robot's arrival defines the effective waiting time at a goal node $j$, the aggregate cost function $\Lambda$ (see Def.~\ref{def:aggregate}) is equivalent to the \emph{minimum} operator. The effective waiting time (cost) at goal $j$ is
\begin{equation}
\Lambda(I_j(\mathcal{A})) \triangleq \min\{ C_{ijk} | (i,j,k) \in I_j(\mathcal{A})\}. \label{eq:min}
\end{equation}
\end{definition}

This application satisfies the conditions in Proposition~\ref{theorem:alg_greedy}: In~\cite{prorok:2018} we show that Eq.~\ref{eq:min} is supermodular. Further, the \emph{minimum} operator is a distributive aggregate function. It follows that the objective of minimizing the average effective waiting time is supermodular. The matroid constraint is trivially satisfied. 

\section{Evaluation}
\label{sec:eval}
\begin{figure}[tb]
\centering
\psfrag{t}[cc][][0.7][90]{Expected Travel Time [s]}
\psfrag{r}[cc][][0.7]{Random}
\psfrag{u}[cc][][0.7]{Repeated Hung.}
\psfrag{e}[cc][][0.7]{Greedy}
\psfrag{C}[cc][][0.7][90]{Correlation}
\subfigure[]{\includegraphics[width=0.43\columnwidth]{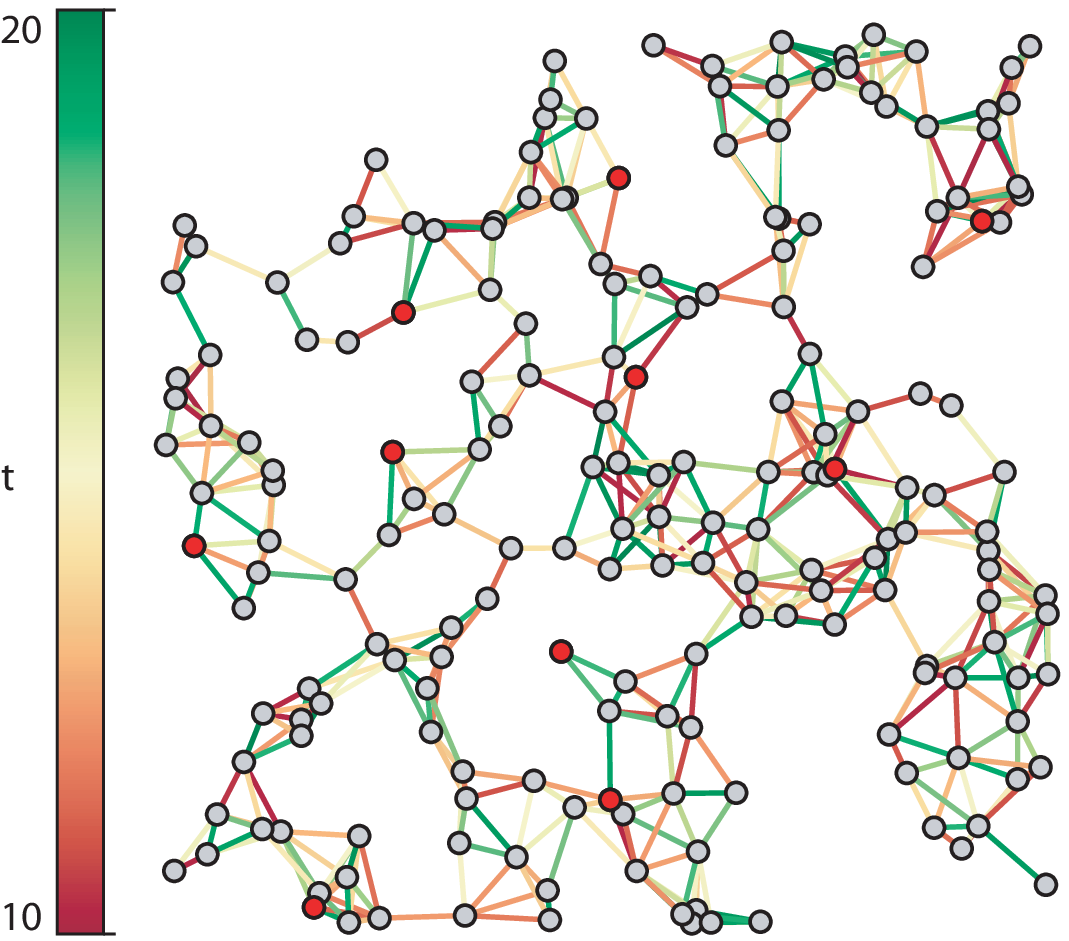}\label{fig:graph_instance_a}}\hspace{0.3cm}
\subfigure[]{\includegraphics[width=0.49\columnwidth]{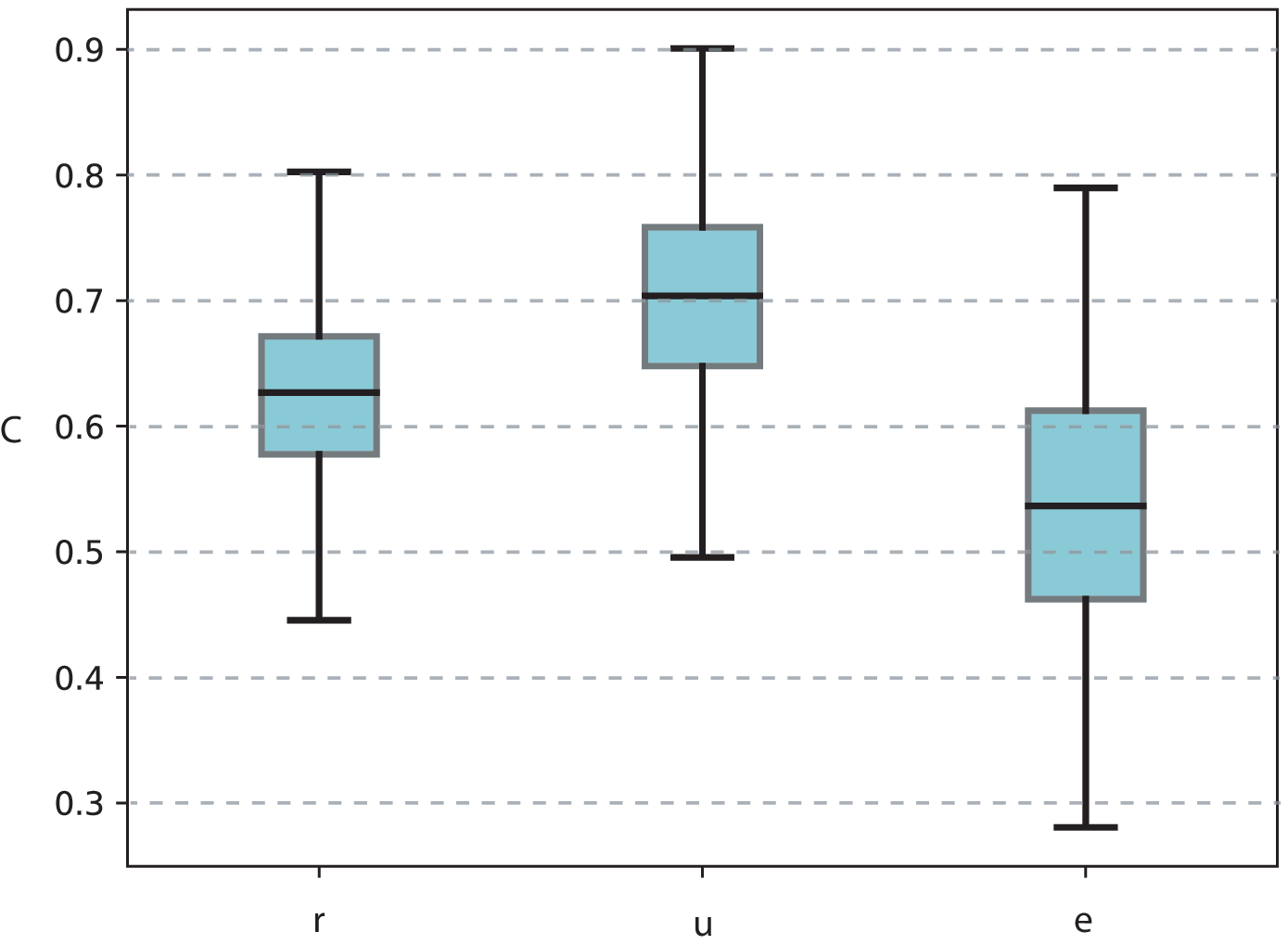}\label{fig:graph_instance_b}}
\caption{(a) Instance of random graph. Red dots indicate robot hubs. The edges are colored to indicate the expected travel time. (b) Comparison between \emph{Greedy}, \emph{Random}, and \emph{Repeated Hungarian}, with respect to the correlation of paths in the solution set. 
\label{fig:graph_instance}}
\end{figure}

We evaluate Algorithm~\ref{alg1} in simulation, on a set of random undirected connected graphs with 200 nodes (of which Fig.~\ref{fig:graph_instance_a} shows an example).
Our default values are $N=25$ robots, $N_{\mathrm{d}}=20$ robots, $M=5$ goals, $S=200$ samples, $K=4$ path options. We generate the $K$ path options by taking the shortest (in average) $K$ paths from $i$ to $j$.
Robots are initially located at 10 randomly selected hubs.
%
The joint distribution of travel times along all transport edges is modeled as a multi-variate Gaussian (truncated at 0) with a mean sampled uniformly at random between 10 and 20. Its covariance matrix is such that the diagonal entries are sampled uniformly between 25 and 100, and the off-diagonal correlation factors are generated using a random lower-triangular matrix corresponding to its Cholesky decomposition. 
Each random transport graph allows us to generate an $MNK$-dimensional sample from the underlying assignment cost distribution $\mathcal{D}$. Using $\mathcal{D}$, we also sample a `true' (observed) travel time for each edge, which we use for our performance evaluation (this value is unknown to all algorithms except \emph{Best a-posteriori}, as described below).
\begin{figure}[tb]
\centering
\psfrag{a}[lc][][0.7]{Hungarian}
\subfigure[]{\includegraphics[width=0.42\columnwidth]{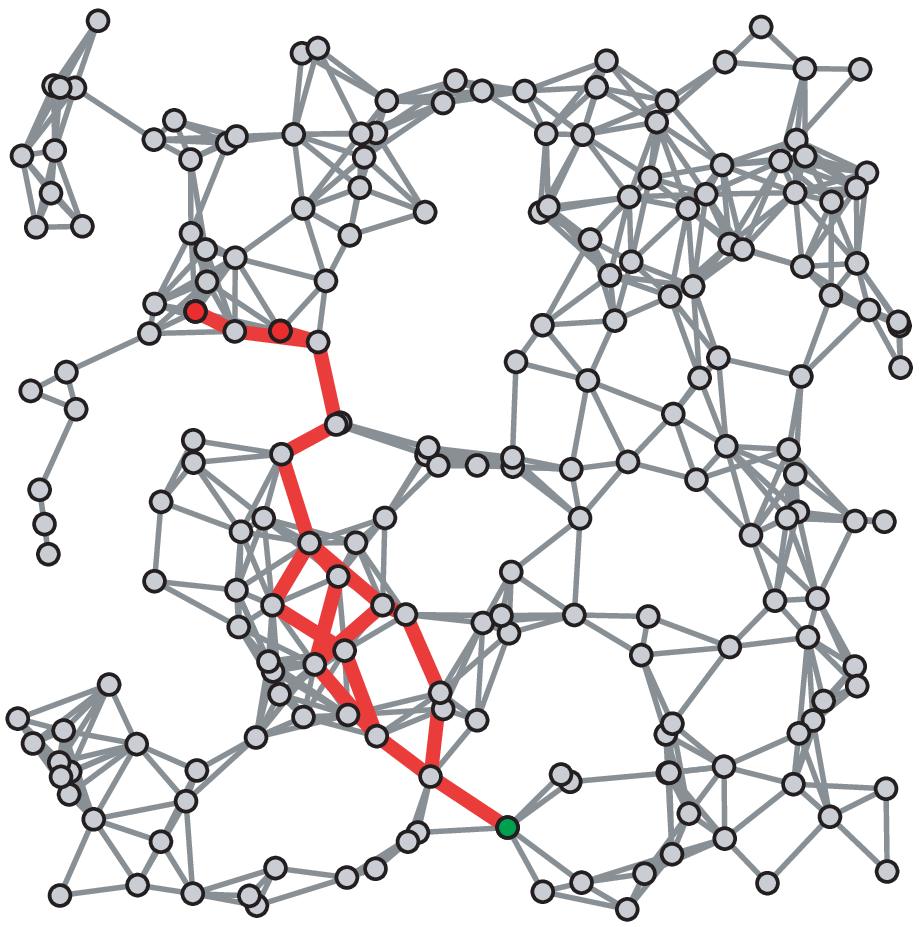}\label{fig:results_paths_a}}\hspace{0.3cm}
\subfigure[]{\includegraphics[width=0.42\columnwidth]{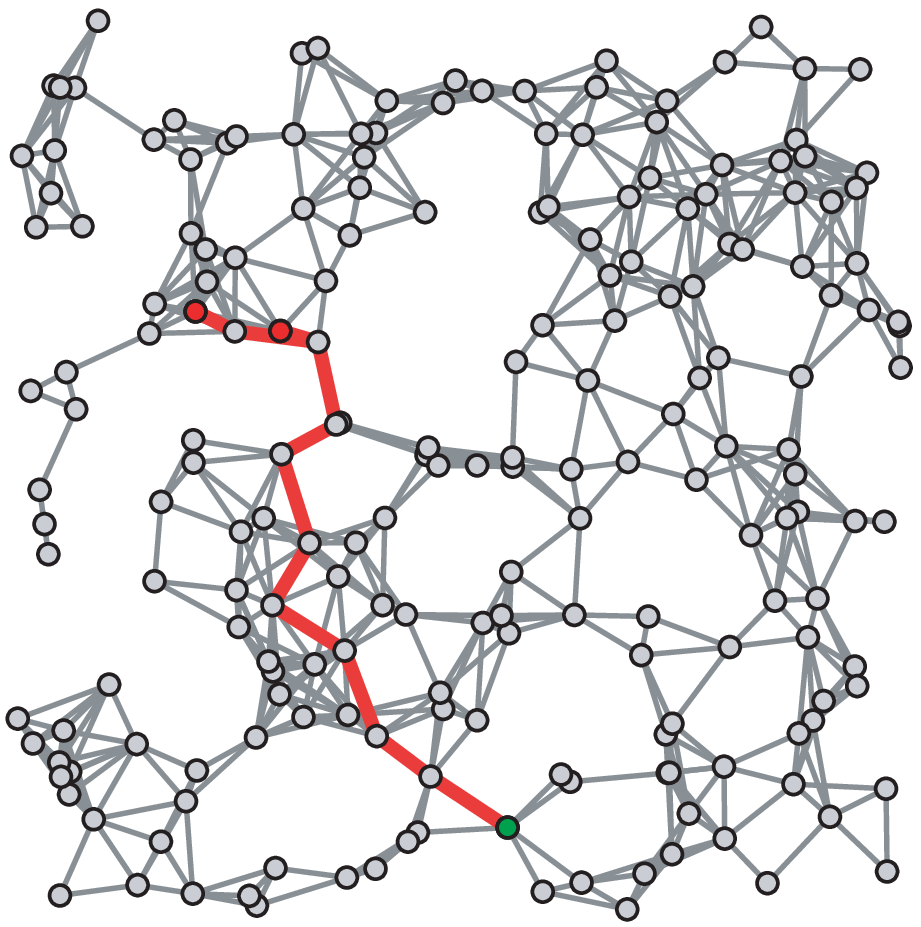}\label{fig:results_paths_b}}
\caption{Paths selected by a robot coalition initially located at two separate hubs (red nodes), and assigned to a goal (green node) for (a) \emph{Greedy} and (b) \emph{Repeated Hungarian}.
\label{fig:results_paths}}
\end{figure}
The performance of our method (\emph{Greedy}) is compared to four alternate assignment algorithms: 
\textbf{(1)} \emph{Hungarian:} We implement the Hungarian method on average waiting times for a non-redundant assignment of $N_{\mathrm{d}}= M$ robots (i.e., $\mathcal{A}=\emptyset)$. This represents the initial assignment $\mathcal{O}$, and is used as the baseline for all following (redundant) assignment algorithms.
\textbf{(2)} \emph{Random:} A random algorithm assigns the redundant $N_{\mathrm{d}} - M$ robots randomly to goals.
\textbf{(3)} \emph{Repeated Hungarian:} We implement repeated iterations of the Hungarian assignment algorithm (at each iteration, assigning $M$ redundant robots in one go), until the cap $N_{\mathrm{d}} - M$ is reached.
\textbf{(4)} \emph{Best a-posteriori:} This corresponds to the best a-posteriori performance for a given set of robot origins and goal destinations, based on true (observed) travel times, on which we run the Hungarian method with $N_{\mathrm{d}}= M$ robots. 
\begin{figure}[tb]
\centering
\psfrag{a}[lc][][0.7]{Hungarian}
\psfrag{e}[lc][][0.7]{Random}
\psfrag{o}[lc][][0.7]{Repeat Hung.}
\psfrag{n}[lc][][0.7]{Greedy}
\psfrag{s}[lc][][0.7]{Best a-post.}
\psfrag{x}[cc][][0.7]{$K$ Paths}
\psfrag{m}[cc][][0.7]{Deployment Size $N_{\mathrm{d}}$}
\psfrag{c}[cc][][0.7][90]{Waiting Time [s]}
\subfigure[]{\includegraphics[width=0.48\columnwidth]{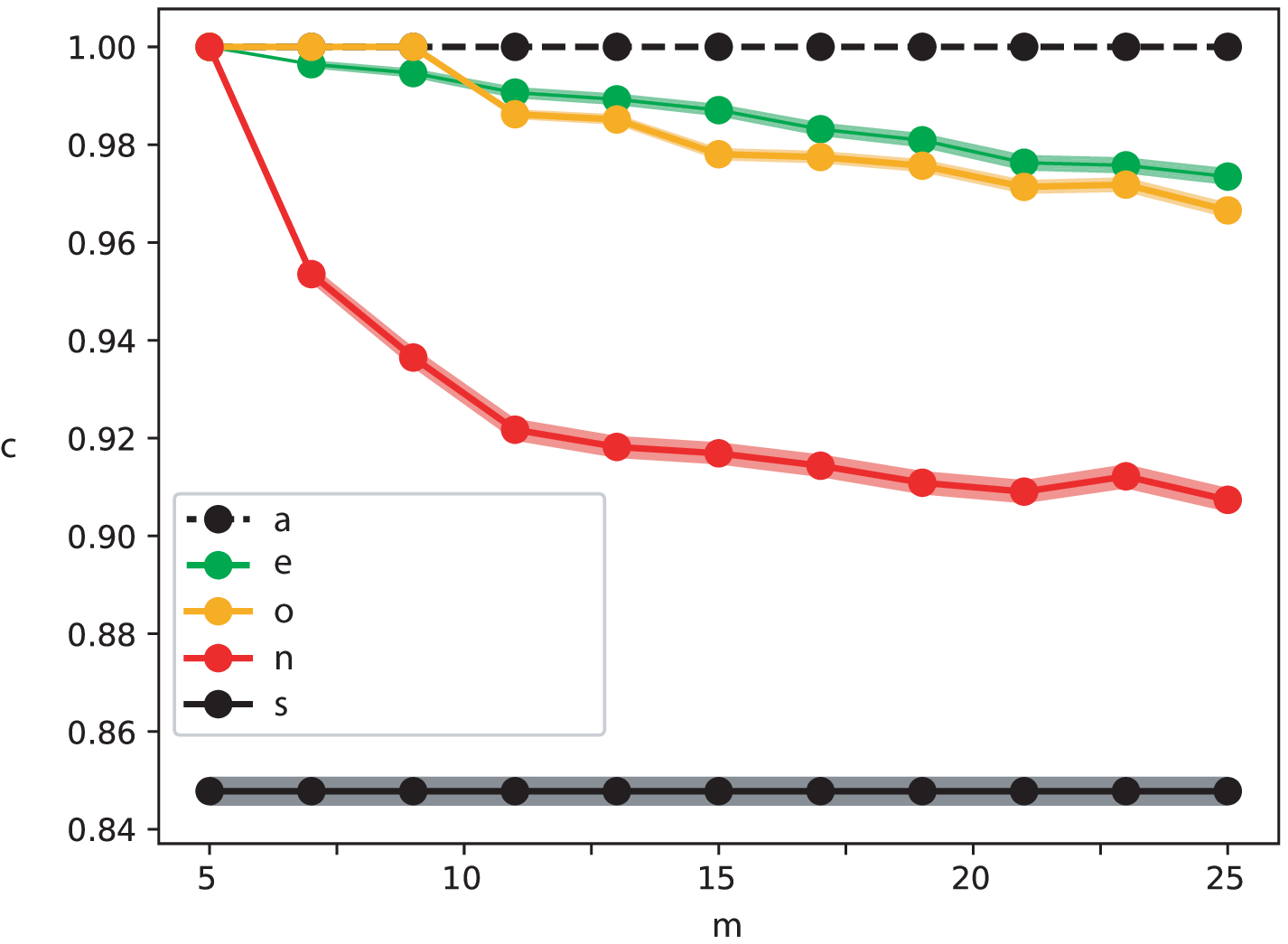}\label{fig:results_time_a}}\hspace{0.3cm}
\subfigure[]{\includegraphics[width=0.48\columnwidth]{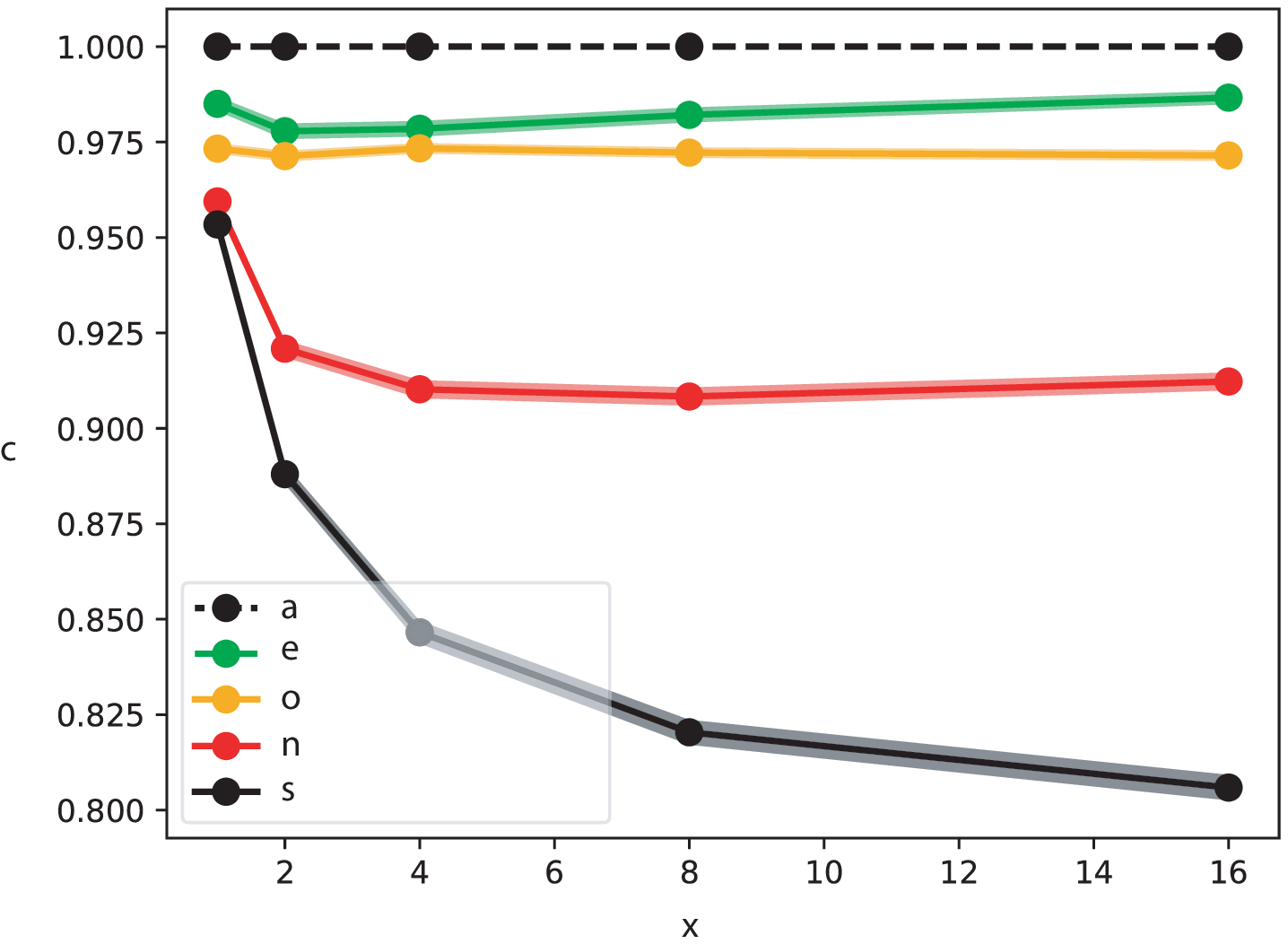}\label{fig:results_time_b}}
\caption{Performance of the assignment strategies, as measured by normalized waiting time. Each data-point is averaged over 500 runs, and the shaded areas show 95\% confidence intervals. (a) Performance as a function of deployment size $N_{\mathrm{d}}$ for $N=25$. (b) Performance as a function of $K$ paths, $N_{\mathrm{d}} = 20$.
\label{fig:results_time}}
\end{figure}

Fig.~\ref{fig:graph_instance_b} shows the correlation of paths in the solutions found by three strategies (\emph{Greedy}, as well as \emph{Repeated Hungarian} and \emph{Random}). 
For each robot coalition assigned to one goal, we compute the average pairwise correlation between all pairs of paths found for the robots belonging to that coalition. The latter value is averaged over all coalitions. 
We observe that the correlation of coalition paths generated by \emph{Greedy} is lower than that of both \emph{Random} and \emph{Repeated Hungarian}.
This indicates that paths selected by \emph{Greedy} tend to be more diverse. This is exemplified in Fig.~\ref{fig:results_paths}, which shows paths selected by a coalition of 5 robots using \emph{Greedy} in Fig.~\ref{fig:results_paths_a} and \emph{Repeated Hungarian} in Fig.~\ref{fig:results_paths_b}.

Fig.~\ref{fig:results_time} shows the performance of our algorithm, as measured by the normalized waiting time $J/J_0$. Fig.~\ref{fig:results_time_a} shows how, as we increase the total robot deployment $N_{\mathrm{d}}$, the waiting time decreases, approaching the lower bound (given best a-posteriori). Fig.~\ref{fig:results_time_b} shows how, as we increase the number of paths $K$ to be considered by the assignment algorithm, performance improves initially, but then flattens out. This validates our usage of a fixed cap ($K$) on the number of paths to be considered by the algorithm. 
Overall, we see that any redundant assignment strategy improves upon non-redundant assignment. 
Our solution \emph{Greedy} improves significantly upon the benchmarks \emph{Random} and \emph{Repeated Hungarian}.

\section{Related Work}

Our problem is related to the general class of submodular welfare problems~\cite{lehmann:2001,vondrak:2008}, which stems from the field of combinatorial auctions. The welfare problem considers a set of items and a set of players, and seeks a partition of the items into disjoint sets assigned to players in order maximize the total welfare over the players. The \emph{welfare} is equivalent to the sum of utilities over all sets. The utility functions satisfy the property of diminishing returns, and hence, the problem can be formulated as one of submodular maximization.
In contrast to our work, the welfare problem does not prescribe any explicit form for the submodular function. Instead, it assumes a \emph{value oracle model}, which is a black-box that returns the utility for any given set. 
In this sense, our problem is a specialization of the general submodular welfare problem. We consider a specific objective function, where the submodularity (or, in our case, the supermodularity) arises due to the redundancy of assigned robots with uncertain travel times.
Furthermore, we also provide a specialization of the matroid constraint, which, in our case, relates to the constraint on maximum possible robot deployment sizes.  

Another related body of work deals with the weapon-target assignment problem~\cite{ahuja:2007}, which considers the assignment of weapons to targets so that the total expected survival value of the targets is minimized. Similar to our problem, this problem considers the assignment of a redundant number of items (i.e., weapons) to a single goal (i.e., target), where assignment costs are uncertain (i.e., probability of target survival). 
In contrast to our work, however, the weapon-target assignment problem only applies to binomial distributions that model the outcome (i.e, survival) of each target as a Bernoulli random variable. Our algorithm is capable of dealing with arbitrary (and potentially correlated) probability distributions that describe the assignment costs.
In that sense, our problem is a generalization of the weapon-target assignment problem.

Submodular optimization for combinatorial problems has has also gained considerable traction in the domain of multi-robot systems. Applications include coordinated robot routing for environmental monitoring~\cite{singh:2009}, leader selection in leader-follower systems~\cite{clark:2014}, sensor scheduling for localization~\cite{singh:2017}, and path planning for orienteering missions~\cite{jorgensen:2017}. Typically, the aforementioned studies develop explicit objective functions that are specific to the considered problem domains. The authors then go about proving the submodularity property and devising the appropriate assignment algorithms. This general methodology is similar to the one presented in our paper.
The particularity of our work lies in the specificity of our objective function for redundant robot assignments under uncertain travel times, which is not captured by objective functions developed in prior work. 

\section{Conclusion} 
\label{sec:conclusion}
In this work, we provided a framework for redundant robot-to-goal assignment. Our main contribution is a supermodular optimization framework that efficiently selects robot matchings and the corresponding paths that minimize the average waiting time at the goal locations when travel times along paths are uncertain.
We introduced a polynomial-time algorithm that uses distributive aggregate functions to efficiently assemble the gain of assigning redundant robots. Our evaluations on random graphs with uncertain and partially correlated edge costs showed that redundant assignment reduces the waiting time at goals with respect to non-redundant assignments. In particular, our proposed algorithm significantly outperforms alternate benchmark algorithms. 
We also demonstrate that our method creates robot coalitions that tend to be more complementary, and thus, more diverse in their selected path options.
In conclusion, we show how providing redundant robots can be a key design feature for systems where time is the primary asset, and where an over-provisioning with respect to the number of robots is a minor concern.



\section*{Acknowledgements}
We gratefully acknowledge the support of {ARL grant DCIST CRA W911NF-17-2-0181.}

\bibliographystyle{spbasic}
\bibliography{Bibliography}

\end{document}